\documentclass[sigconf]{acmart}
\usepackage{booktabs} % For formal tables
\usepackage{amsmath}
\usepackage{algorithm}
\usepackage[noend]{algpseudocode}
\usepackage{xspace}
\usepackage{enumitem}
\usepackage{color, colortbl}
\usepackage{hhline}

\definecolor{Gray}{gray}{0.8}

\newcommand{\MethodName}{PISA\xspace}
\def\Unanonymous{1}

\ifx\Unanonymous\undefined

\else

\fi

\setcopyright{rightsretained}
\acmDOI{10.475/123_4}

\acmISBN{123-4567-24-567/08/06}

\editor{Jennifer B. Sartor}
\editor{Theo D'Hondt}
\editor{Wolfgang De Meuter}

\begin{document}
\title {New Item Consumption Prediction Using Deep Learning}
%\title{Addressing Cold Start in Consumption Intent Prediction for Sessions using Deep Learning}
%\titlenote{Produces the permission block, and   copyright information}
%\subtitle{Extended Abstract}
%\subtitlenote{The full version of the author's guide is available as
%  \texttt{acmart.pdf} document}

\ifx\Unanonymous\undefined

\author{Anonymous}
%\authornote{Dr.~Trovato insisted his name be first.}
\orcid{}
\affiliation{%
	\institution{Anonymous}
	%\streetaddress{P.O. Box 1212}
	\city{}
	\state{}
	\postcode{}
}
\email{}

\else
\author{Michael Shekasta}
%\authornote{Dr.~Trovato insisted his name be first.}
\orcid{1234-5678-9012}
\affiliation{%
	\institution{Ben-Gurion University of the Negev}
	%\streetaddress{P.O. Box 1212}
	\city{Beer Sheva}
	\state{Israel}
	\postcode{43017-6221}
}
\email{shkasta@post.bgu.ac.il}
\author{Gilad Katz}
%\authornote{Dr.~Trovato insisted his name be first.}
\orcid{1234-5678-9012}
\affiliation{%
  \institution{Ben-Gurion University of	the Negev}
  %\streetaddress{P.O. Box 1212}
  \city{Beer Sheva}
  \state{Israel}
  \postcode{43017-6221}
}
\email{katzgila@post.bgu.ac.il}

\author{Asnat Greenstein-Messica}
%\authornote{Dr.~Trovato insisted his name be first.}
\orcid{1234-5678-9012}
\affiliation{%
	\institution{Ben-Gurion University of	the Negev}
	%\streetaddress{P.O. Box 1212}
	\city{Beer Sheva}
	\state{Israel}
	\postcode{43017-6221}
}
\email{asnatm@post.bgu.ac.il}

\author{Lior Rokach}
%\authornote{Dr.~Trovato insisted his name be first.}
\orcid{1234-5678-9012}
\affiliation{%
	\institution{Ben-Gurion University of	the Negev}
	%\streetaddress{P.O. Box 1212}
	\city{Beer Sheva}
	\state{Israel}
	\postcode{43017-6221}
}
\email{liorrk@bgu.ac.il}

\author{Bracha Shapira}
%\authornote{Dr.~Trovato insisted his name be first.}
\orcid{1234-5678-9012}
\affiliation{%
	\institution{Ben-Gurion University of	the Negev}
	%\streetaddress{P.O. Box 1212}
	\city{Beer Sheva}
	\state{Israel}
	\postcode{43017-6221}
}
\email{bshapira@post.bgu.ac.il}

\fi
\iffalse
\author{G.K.M. Tobin}
\authornote{The secretary disavows any knowledge of this author's actions.}
\affiliation{%
  \institution{Institute for Clarity in Documentation}
  \streetaddress{P.O. Box 1212}
  \city{Dublin}
  \state{Ohio}
  \postcode{43017-6221}
}
\email{webmaster@marysville-ohio.com}

\author{Lars Th{\o}rv{\"a}ld}
\authornote{This author is the
  one who did all the really hard work.}
\affiliation{%
  \institution{The Th{\o}rv{\"a}ld Group}
  \streetaddress{1 Th{\o}rv{\"a}ld Circle}
  \city{Hekla}
  \country{Iceland}}
\email{larst@affiliation.org}

\fi
\renewcommand{\shortauthors}{Shekasta et al.}

\begin{abstract}
Recommendation systems have become ubiquitous in today's online world and are an integral part of practically every e-commerce platform. While traditional recommender systems use customer history, this approach is not feasible in 'cold start' scenarios. Such scenarios include the need to produce recommendations for new or unregistered users and the introduction of new items. In this study, we present the Purchase Intent Session-bAsed (\MethodName) algorithm, a content-based algorithm for predicting the purchase intent for cold start session-based scenarios. Our approach employs deep learning techniques both for modeling the content and purchase intent prediction. Our experiments show that PISA outperforms a well-known deep learning baseline when new items are introduced. In addition, while content-based approaches often fail to perform well in highly imbalanced datasets, our approach successfully handles such cases. Finally, our experiments show that combining \MethodName with the baseline in non-cold start scenarios further improves performance.
\end{abstract}



\maketitle

\section{Introduction}

Recommendation systems (RSs) \cite{bobadilla2013recommender} aid users in handling information overload by recommending unfamiliar items that suit the user's preferences and needs. RSs collect information on the user's preferences for the items in a given domain (e.g., music, movies, e-commerce, etc.) and then attempt to predict what other items the user is likely to find relevant. Information on users' preferences may be acquired explicitly (e.g., with rating, like/dislike, etc.) or by implicitly monitoring the users' actions.

The most popular recommender system approaches include the \cite{adomavicius2005toward}: \textit{collaborative}, \textit{content-based}, and \textit{hybrid} techniques. Collaborative filtering algorithms \cite{ricci2015recommender, herlocker2017algorithmic}  rely on the similarity of collaborative historical data both of users and items to produce future recommendations. The content-based approach \cite{balabanovic1997fab,pazzani1999framework,pazzani2007content} attempts to recommend items based on the similarity of their content features, such as description (e.g., an item catalog) or proximity in a taxonomy. Hybrid recommendation algorithms \cite{burke2002hybrid} often combine the approaches presented above to create a more robust recommendation model.

In this study, we focus on session-based recommendations \cite{ludewig2018evaluation}. In this setting, we attempt to determine whether a sequence of items reviewed by the user during a session is likely to end with a purchase. In addition to being reflective of real-world scenarios, this problem is also challenging because of the need to model inter-item dependencies. Recent studies related to session-based recommendation  \cite{hidasi2017recurrent,hidasi2015session,greenstein2017session} 
focus on predicting the next item of the session and producing a top-k recommendation list, rather than predicting the consumption intent. We believe that predicting the intent of the user early in the session might open the door to numerous methods aimed at improving the session outcome. One example is the application of intervention during a session based on the predicted intent. Thus, if for example, the system predicts that a user is likely to leave the session without buying, it might offer a discount to change the user's intent. Moreover, we also address the cold start scenario for purchase intent prediction in sessions where the history of the user's purchasing behavior is not available for new items in the system. This scenario is typical to dynamic e-commerce sites that add new items regularly to the inventory. Unfortunately this challenge has not yet been addressed by the research community.

We present \MethodName, a content-based purchase prediction method for session-based recommendations. Our approach consists of two phases: first, we use the item descriptions and categories to create word embeddings that model the relationships among items. Next, we use these embeddings to model the items in each session and predict the likelihood of a purchase. We evaluate the performance of our approach on a large commercial dataset containing over 1.6M sessions and 18,000 items.

Our experiments show that the proposed approach can significantly outperform the deep learning state of the art baseline in cold start scenarios, which are considered to be one of the main challenges to recommendation systems \cite{schein2002methods}. In addition, we show that when we integrate the proposed approach with existing baselines, we obtain greater improvement. Finally, our evaluation shows that \MethodName performs well on highly imbalanced datasets, a setting that is very difficult for recommendation systems \cite{ben2015recsys,romov2015recsys}.

Our contributions in this study are as follows:

\begin{itemize}
	\item We present a novel content-based purchase prediction approach for session-based purchase prediction. Our approach utilizes word embeddings to model the relations among items and can be used both on its own and in combination with standard collaborative filtering approaches.
	\item We demonstrate the effectiveness of our approach in cold start scenarios and in cases where the data is highly imbalanced. These two scenarios are known to be particularly challenging for most recommendation algorithms.
	\item We evaluate our results on a large commercial dataset to validate our results and analyze the performance of our approach on different product categories.
\end{itemize} 

\section{Related work}
\subsection {Session-Based Recommendation} 
Much of the work in the area of recommender systems has focused on models that work when a user is identified in a system and a clear user profile can be built. Session-based recommendation, where a user is anonymous or not logged in yet and the recommendations are based on short session available data instead of a lengthy user history, is quite common in real-life. In such cases, the item-to-item recommendation approach is commonly used \cite{jannach2017session}. Items which are usually clicked or bought along with the items the user clicked are recommended. Another approach used for session-based recommendation
\cite{ludewig2018evaluation} uses Markov Decision Process (MDP) methods, which are based on sequential stochastic decision problems and Bayesian Personalized Ranking \cite{rendle2009bpr}. Recently, recurrent neural networks (RNNs) were applied to session-based recommendation with excellent results \cite{hidasi2015session}. The sequence of items the user clicked during the session is fed into the RNN to predict other items the user may like. Since representing items by one-hot encoding drastically increases the feature space because of the large item inventory, dimensionality reduction using Word2Vec \cite{mikolov2013efficient} or GloVe \cite{pennington2014glove} models is often used before feeding the sequence into the RNN \cite{greenstein2017session}. In \cite{hidasi2017recurrent}, the authors showed that incorporating the item's image feature vector into the RNN further improves the accuracy of the recommendation.

One of the challenges associated with session-based recommendation is predicting the consumer's consumption intent based on the user clicks so far
\cite{ben2015recsys}. Based on this prediction, an e-commerce vendor may suggest different promotions to the consumer to improve the conversion rate. One of the key challenges of this task stems from the extreme class imbalance of the data, since only a small fraction of the sessions conclude with a purchase. Gradient boosting trees combined with an : intensive feature engineering  was used by \cite{romov2015recsys} to solve the consumption intent in the 2015 RecSys Challenge \cite{ben2015recsys}. In \cite{bogina2016learning}, the researchers show that using temporal dynamic features is effective for this purpose. Recently, \cite{ravindran2018neural}  used a combination of a rich set of session-based features and a clickstream representation of the session to predict the consumption intent.
\subsection {Item Cold Start Problem}
The cold start problem is one of the major challenges in the design and deployment of recommender systems. An item cold start occurs when new items are introduced into the system. In e-commerce scenarios, new items are constantly added and hence, there is a need to address the item cold start problem. In this situation, the historical behavioral data (ratings, purchases, clicks, etc.) required for the item-to-item based approach to work properly is lacking. 

Several methods have been introduced to address this problem \cite{schein2002methods}. Most of the proposed methods adopt a content-based approach and utilize the content of new items, in order to identify items with similar content, and subsequently recommend these new items to users. Recent research leveraged a deep learning approach to generate an embedded item representation to address the item cold start problem in a collaborative filtering scenario, where new items should be recommended to recurrent users. In \cite{bansal2016ask}, the authors showed that using an embedded text representation based on an RNN for rich text items, such as scientific paper recommendation, outperforms the state of the art matrix factorization approach for the item cold start scenario. Furthermore, \cite{vartak2017meta} present a meta-learning strategy implemented by a deep neural network architecture to address item cold start for tweet recommendation. Their approach significantly beats the matrix factorization approach for this scenario.

Unlike previous research which leveraged a deep neural network-based model to address the item cold start scenario in a collaborative filtering scenario where new items are recommended to recurrent users, the proposed PISA approach leverages deep neural network architecture to address the item cold start problem to predict highly imbalanced consumption intent for anonymous users, where no historical user behavioral data is available.

\section{Problem Formulation}
The underlying assumption in this research is that users commonly examine several items prior to their decision to purchase. Hence, we model users' session activity as a sequence of click events for items and purchase events. For example, (c1, c2, c3, c4, b4) denotes a user session consisting of four click events on four different items followed by a single buy event. Our goal is to predict whether the user will purchase at least one item during a given session (i.e., consumption intent), based on the user's first few clicks. By learning to predict the consumer's consumption intent, one can improve the overall sales conversion by selectively offering promotions.\\

\noindent \textbf{General definitions.}  For our task, we assume that each user is capable of carrying out the following actions on a set of items $I$ which belong to the product catalog:
\begin{itemize}
	\item ``Buy'' -- purchase item $i \in I$. Denoted by $b_{i}$ 
	\item ``Click'' --  click on item $i \in I$. Denoted by $c_{i}$
\end{itemize}
A session $S_j$ of length $L$ is defined as a sequence of $L$ click events that a specific user performed in an e-commerce website on different items ${i_m} \in I$ within a time window of 24 hours
$(c_{j,i_1},c_{j,i_2},...,c_{j,i_L})$. Different sessions are different lengths. Our goal is to predict the probability that the session $j$ will end with a purchase of at least one product (session output $O_{j} = 1$) based on the session's click events $P(O_{j}=1|S_j)$.
\noindent \newline \textbf{Catalog for Content-Based Recommendation.} \noindent We assume the existence of a catalog. For the purpose of this research, the catalog contains the following information about each item $i \in I$:
\begin{itemize}
\item Item category: $C_i \in {C_1,...C_K}$, where $K$ is the number of item categories
\item Item description: $D_i$, which is represented by a sequence of $Q$ words ($(w_{i1}..w_{iQ})$) where each word $w_{id}$ elongs to a predefined vocabulary $V$. The description length $Q$ may vary among items. Each word is represented by a one-hot vector $h$, which contains $V$ bits. The bit which corresponds to the specific word index in the vocabulary is set to one, while the rest of the bits are set to zero.
\item Item title: $T_i$, which is represented by a sequence of $R$ words $(w_{i1}..w_{iR})$ where each word $w_{it}$ belongs to a predefined vocabulary. The title length $R$ may vary among items. Each word is represented by a one-hot vector of size $V$ as described above.
\end{itemize}
\section{The Proposed Method}
\label{sec:proposedMethod}

\noindent We present a deep learning content-based algorithm that utilizes the description of items (from a catalog) and the items' categories to enhance the ability to predict purchase intent. We hypothesize that the content-based approach is beneficial for cold start scenarios involving new items. We present two variants of our proposed approach: the first relies solely on the textual content (the categories and description of each item), while the second combines the textual approach with the common item ID-based approach. We denote these two approaches as \textit{content-based} and \textit{integrated}, respectively. \\ 

\begin{figure}[H]
	\includegraphics[height=2.25in, width=1.5in]{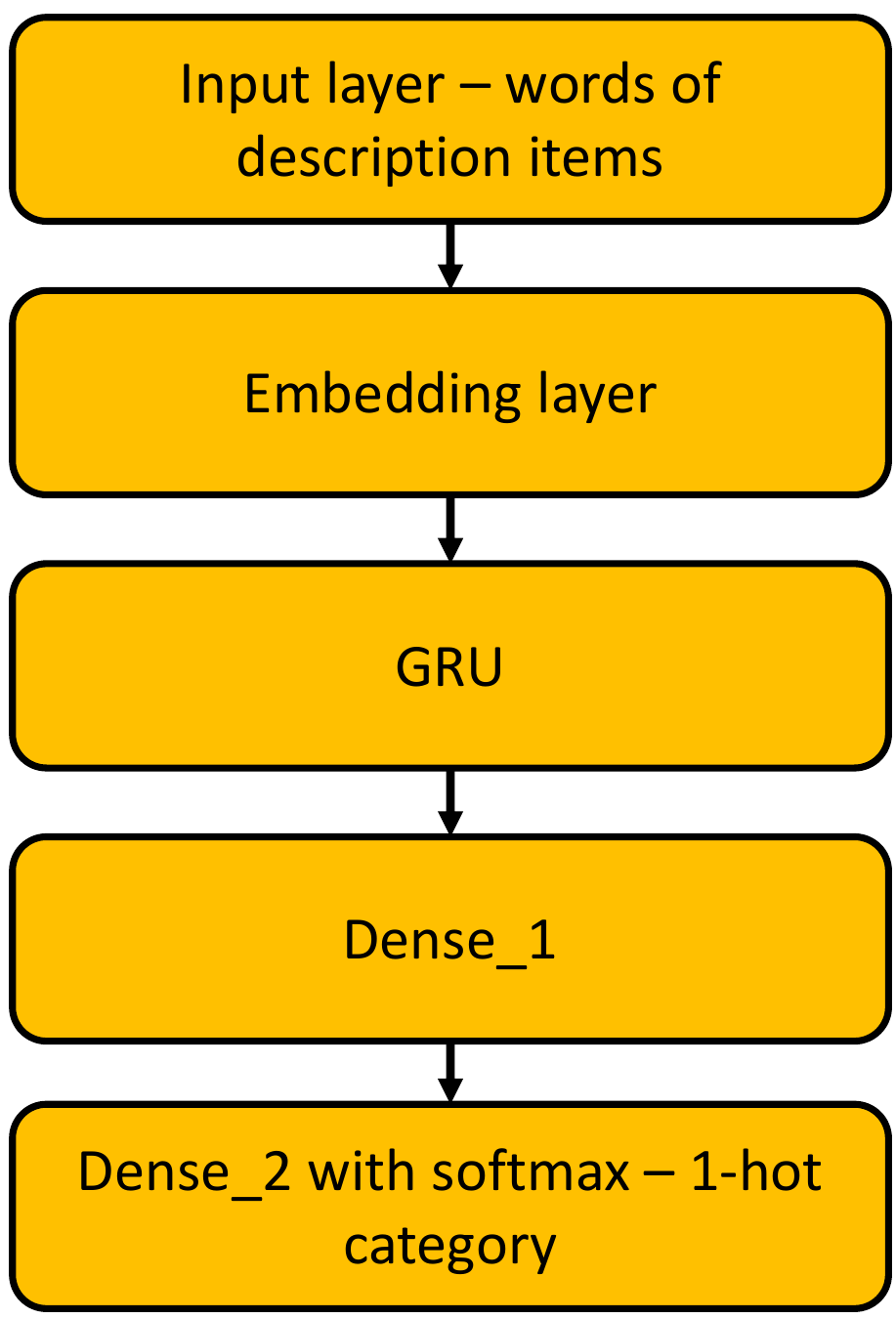}
	\caption{Text model diagram - component one}
	\label{fig:textmodel_diagram}
\end{figure}

\begin{figure}[H]
	\includegraphics[height=1.75in, width=3in]{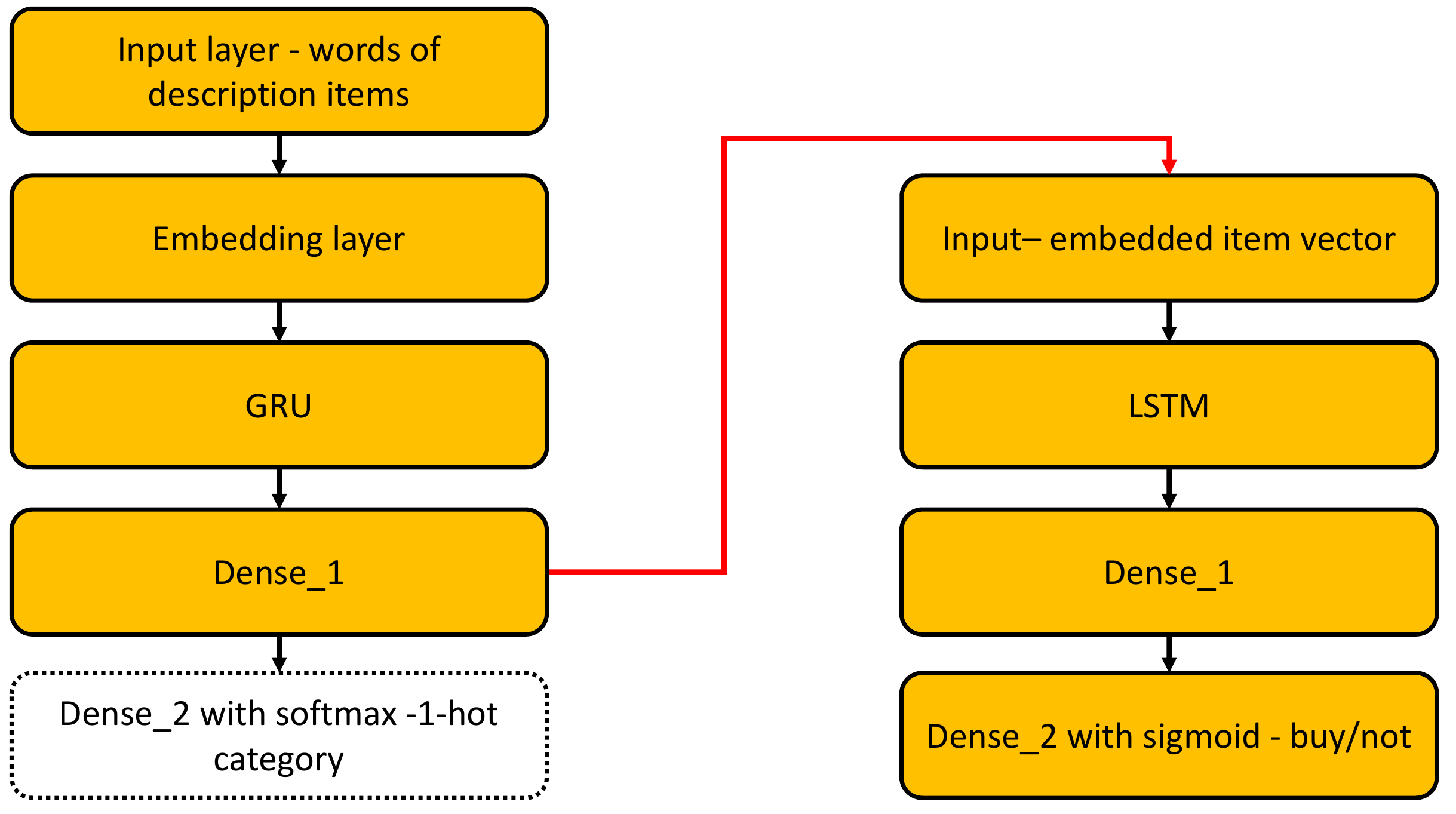}
	\caption{Text model diagram - component two}
	\label{fig:textmodel_diagram_b}
\end{figure}

\noindent \textbf{Content-based Approach (\MethodName).} This approach consists of two components, which are presented in Figures \ref{fig:textmodel_diagram} and \ref{fig:textmodel_diagram_b}; the first component is for generating the item embedding (the embedding component), and the second is for sequence purchase prediction (the prediction component). The output of the first component serves as the input of the second component. We now describe each component in detail.

The goal of the embedding component is to create a dense semantic representation of the items and the categories they belong to. As shown in multiple domains \cite{covington2016deep, bansal2016ask, hu2017diversifying}, embedding-based representations are capable of capturing latent connections among items with little or no explicit correlation. The embedding component receives the description of an item as input and attempts to predict its category, thus encouraging the final embedding to group words that are common to items of the same category closer together. A gated recurrent unit (GRU) layer \cite{chung2014empirical} is added, in order to consider the order of words that appear in the description, as well as the words themselves. Upon completion of the training, we remove the softmax layer \cite{tang2013deep} and use the output of the fully connected layer (``Dense\_1'' in Figure 1) as input for the prediction component. rest of the network remains unchanged (i.e., no further updating of the weights) throughout the remainder of the prediction component.
The prediction component receives a sequence of items as input, with the representation of each component generated by the embedding component. The items in the sequence are analyzed iteratively. Once the entire sequence has been analyzed, the prediction component attempts to predict whether or not a purchase has taken place. While the embedding component utilizes the GRU architecture to generate the embedding, the prediction component utilizes LSTM. This decision, as with the layer dimensions, was made empirically.

The prediction component receives a sequence of items as input, with the representation of each component generated by the embedding component. The items in the sequence are analyzed iteratively. Once the entire sequence is analyzed, the prediction component attempts to predict whether or not a purchase will take place. While the embedding component utilizes the GRU architecture to generate the embedding, the prediction component utilizes LSTM. This decision, as with the layer dimensions, was made empirically.

\begin{figure}[H]
	\includegraphics[height=1.75in, width=3in]{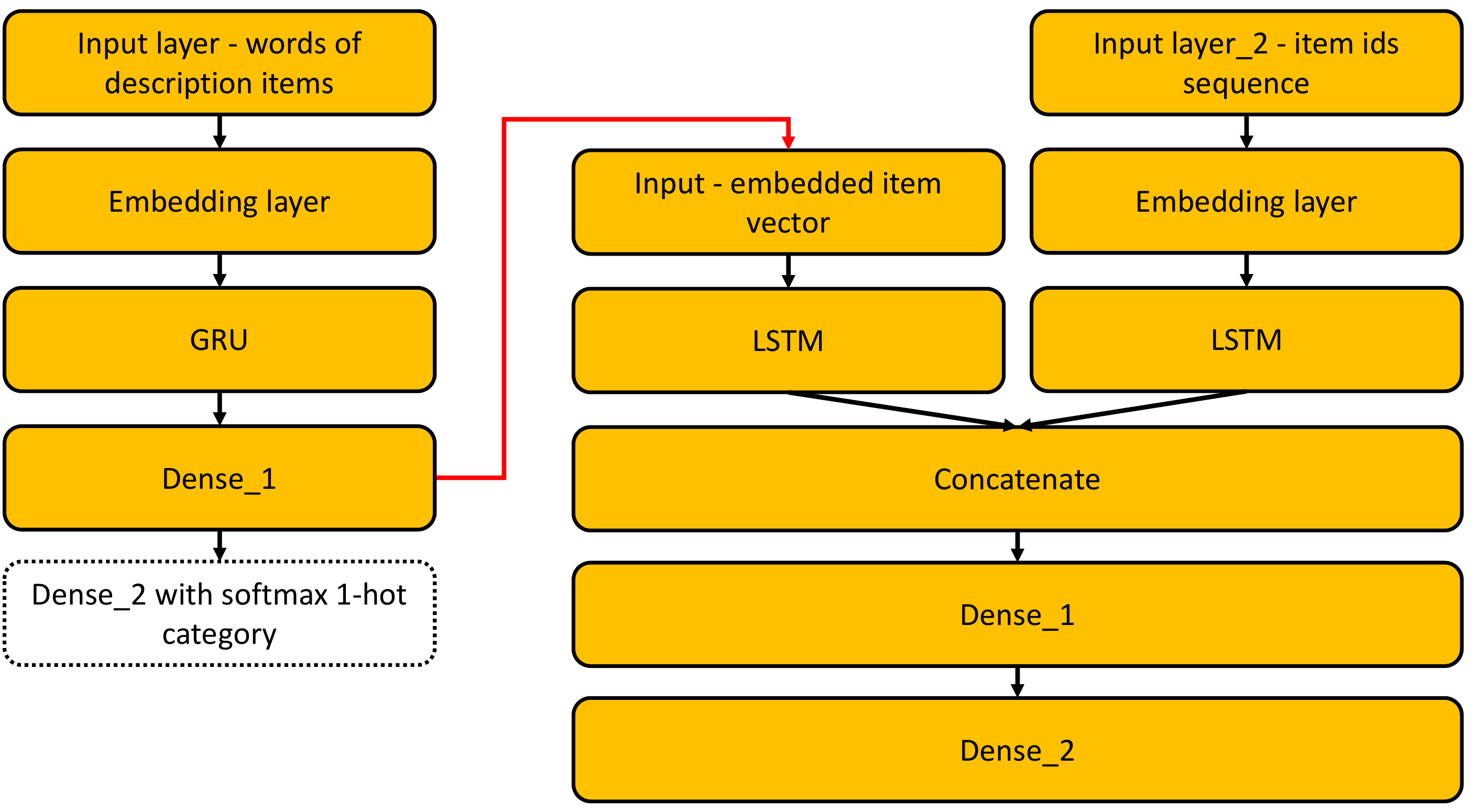}
	\caption{Integrated model diagram}
	\label{fig:integrated_model}
\end{figure}

\noindent \textbf{Integrated approach.} This approach, which is presented in Figure \ref{fig:integrated_model}, consists of the same embedding component described in the previous approach, but the prediction component has been modified so it also combines a sequence of item IDs as input.

The prediction component receives two types of input:
\begin{enumerate}
	\item The sequence of clicked items (the same input as in the content-based approach).
	\item A  sequence of item IDs, along with the ID of the user whose click sequence is currently being analyzed. Each input is analyzed separately using two LSTM layers, and the two outputs are then concatenated and fed into a dense layer (“Dense\_1”). Finally, upon the completion of each sequence, the architecture outputs the likelihood of a purchase.
\end{enumerate}

\section{Evaluation}

\noindent 
\subsection{The Dataset}

\noindent We use a proprietary dataset from a leading e-commerce site that provides personal recommendations of items for registered and guest users. The data consists of \textit{events} and an \textit{item catalog}. This dataset is unique because it includes content data for items (for example, the item description) and event data (clickstream, purchased items).

\noindent \textbf{Events.} We consider the following user actions as events:

\begin{enumerate}
\item  Buy -- the user purchases a specific item.

\item  Click -- the user clicks on an item. 
\end{enumerate}

\noindent For each event, we record the following features: type (buy or click), timestamp, user ID, and item ID.
	
\begin{figure}[H]
	\includegraphics[height=0.25in, width=2.8in]{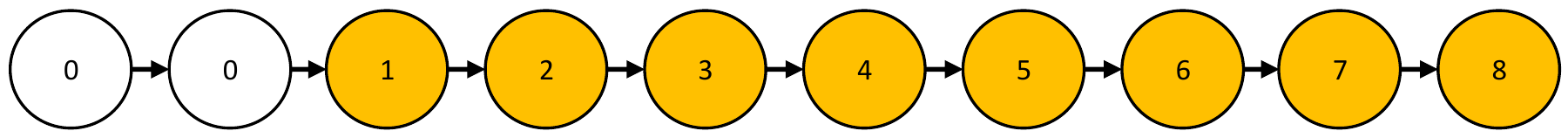}
	\caption{Padding sequence}
	\label{fig:padding}
\end{figure}

\begin{figure}[H]
	\includegraphics[height=0.22in, width=2.86in]{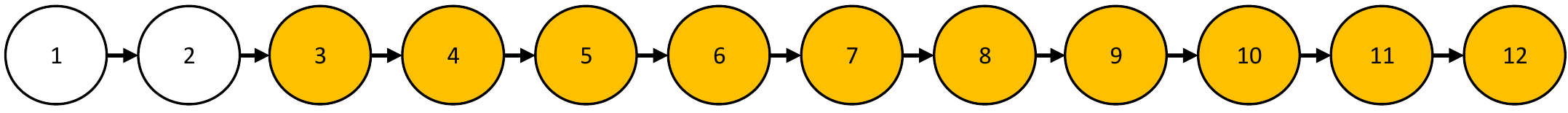}
	\caption{Pruning sequence}
	\label{fig:pruning}
\end{figure}

\begin{figure}[H]
\includegraphics[height=2.in,width=3.2in]{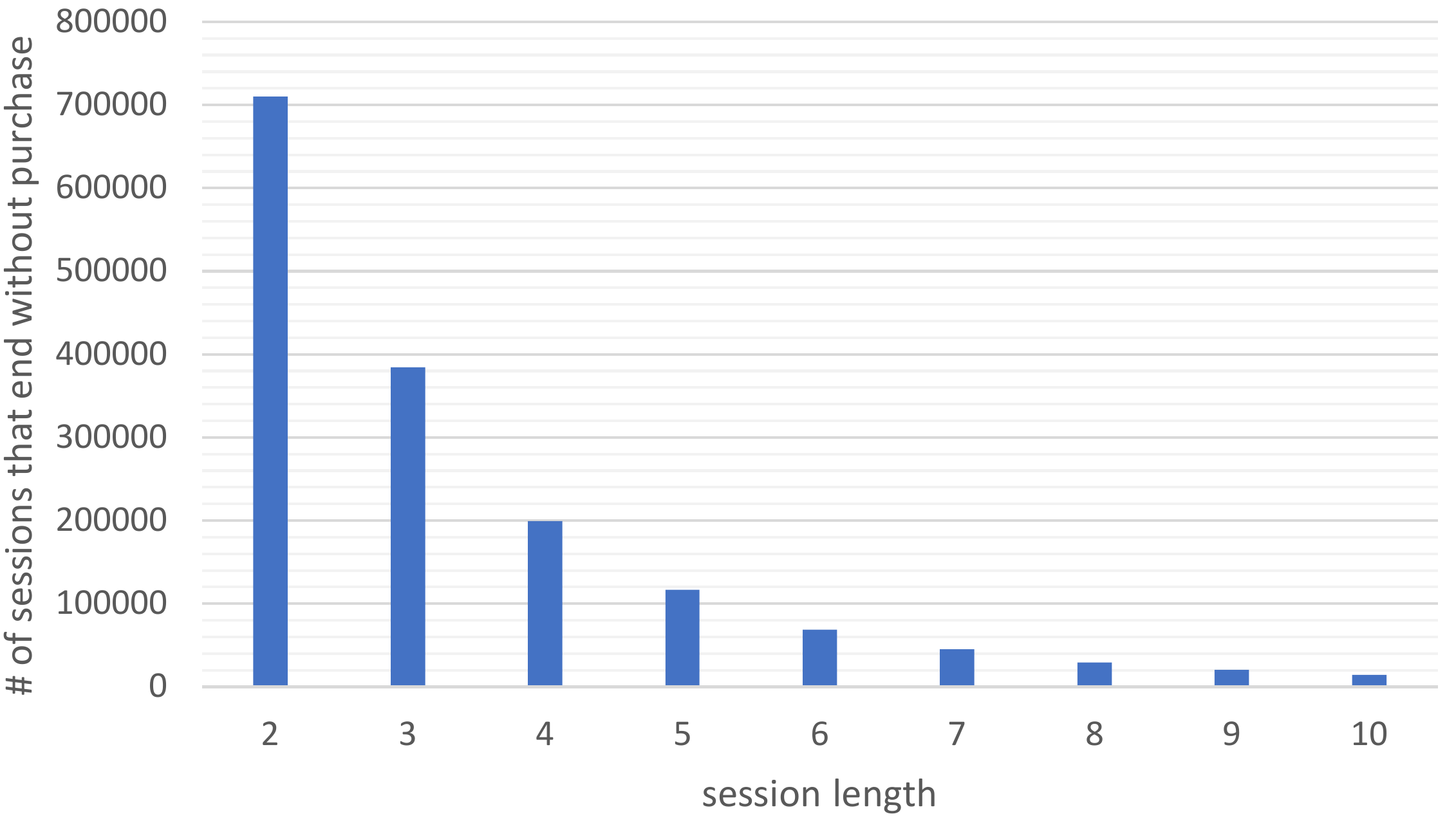}
\caption{The number of sessions of each length and the percentage of sessions that have at least one purchased item}
\label{fig:distribution_length_sessions_no_purchase}
\end{figure}

\begin{figure}[H]
	\includegraphics[height=2.3in, width=3.2in]{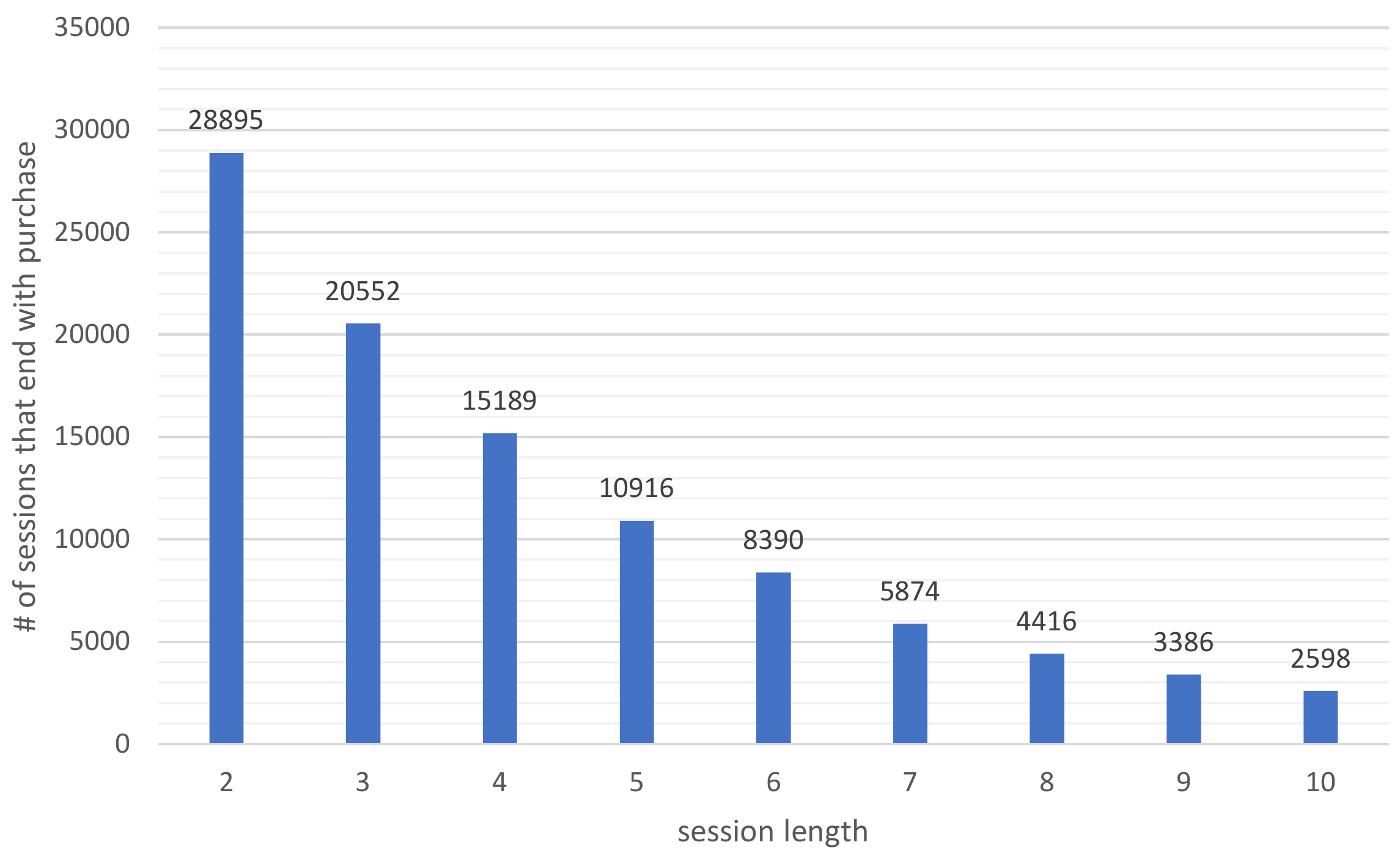}
	\caption{The number of sessions of each length and the percentage of sessions where no item was purchased}
	\label{fig:distribution_length_sessions_purchase}
\end{figure}

Our data was collected during a period of one month and consists of 1,674,963 sessions, 1,505,789 users, 18,308 unique items, 6,471,816 click actions, and 207,438 purchase actions. Approximately 4\% of all sessions end with purchases, where the average number of purchased items per session is three. We define a session as all actions that a user took on the website for a period of 24 hours. We limited our experiments to sessions of up to ten clicked items, in order to filter out very lengthy sessions that are suspected as errors in the data. Figures \ref{fig:distribution_length_sessions_no_purchase} and \ref{fig:distribution_length_sessions_purchase} present the distribution of session length that ends with/without purchase.
To facilitate session analysis, we use \textit{padding} and \textit{pruning} to fit short and long sessions respectively. Examples of these two actions are shown in Figures \ref{fig:padding} and \ref{fig:pruning}. When padding is required, it is added prior to the original events of the session. When pruning is required, we keep the ten most recent events. \newline

\noindent \textbf{The item catalog.} The catalog is written in German. The fields that we consider in our experiments are:

\begin{enumerate}
\item  Item id -- a unique identifier

\item  Item category -- items are divided into 13 categories (books, house \& garden, etc.)

\item  Title -- the name of the product

\item  Short description -- short description in English (roughly one sentence)

\end{enumerate}

The item title and description are appended to create the text describing each item. As explained in Section \ref{sec:proposedMethod}, we use the item category to train the loss function of the embedding component. For this reason, the item category is not directly included in the description of the items.

\subsection{Experimental Setup}
Our data was collected during 30 days in August 2016 (August 31st was not included). We used the data of August 29 and 30 as validation and test sets respectively (as done in \cite{hidasi2015session}) for all scenarios. The number of sessions in the training set was 1,604,640, while the number of sessions in the validation and test sets were 70,323 and 67,753 respectively. 
\label{subsec:experimentalSetup}

We conduct three sets of experiments: \textit{All-Data}, \textit{cold-start}, and \textit{random removal}:

\begin{itemize}
\item \textbf{All-Data.} In this experiment, we run on the data ``as-is''.
\item \textbf{Cold-start.}  We define a cold start session as one that has at least one previously unseen item. However, there are not enough sessions in the test set that meet this criterion for a meaningful evaluation. To increase the number of cold start sessions in the test set, we randomly sample X\% of the items associated with each category from the training set and then remove every session that contains any of the sampled items from the training set.
\item \textbf{Random removal.} This experiment was designed to rule out the possibility that the superior performance of \MethodName (compared to the baseline) in the cold start experiment stemmed from the smaller available dataset rather than its ability to gain new insight from item descriptions. In this experimental setting, we remove the same number of sessions as in the cold start experiments, but we do so randomly.
\end{itemize}

The following settings were used in all experiments (we set the different parameters empirically):
\begin{itemize}
\item The deep neural models used in our experiments were implemented using the Keras library \cite{keras}.
\item We use the nltk library \cite{nltk} to tokenize item descriptions.
\item The baseline used in our experiments was implemented using the Item2Vec model \cite{hidasi2015session}which was trained maximum 20 epochs. The Item2Vec model was also used as the "standard" user-item collaborative recommendation component in the integrated approach presented in Section \ref{sec:proposedMethod}. 
\item The Adam optimizer \cite{kingma2014adam} was used in all of our experiments, along with a learning rate of 0.001.
\item All LSTM and GRU architectures in our experiments consisted of 150 units.
\item Our proposed models were trained for 20 epochs. The validation set is then used to select the top-performing model configuration.
\item We used the AUC measure \cite{hand2001simple} to evaluate the results of our experiments. Our reason for choosing this measure is that it captures algorithmic performance across a range of true positive/false positive rates rather than a single threshold.
\item We used the DeLong statistical test \cite{delong1988comparing} to determine whether the differences in performance among the three algorithms were significant.
\item We also use average precision \cite{zhu2004recall} as an evaluation metric.
\end{itemize}

\subsection{Results}
\noindent We begin by presenting the evaluation results for the three scenarios described in Section \ref{subsec:experimentalSetup}. \newline

\begin{figure}
\includegraphics[height=2.3in, width=3in]{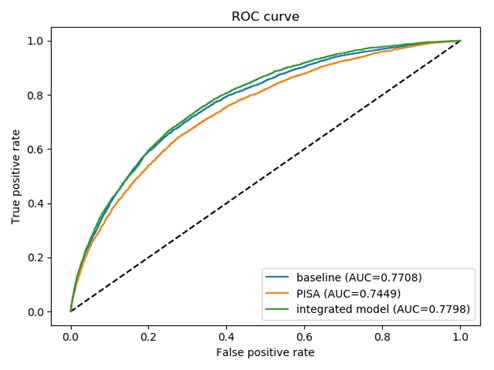}
\caption{Consumption intent prediction - ROC curve}
\label{fig:roc_regular_exp}
\end{figure}

\noindent \textbf{All-Data experiments.} Figure \ref{fig:roc_regular_exp} presents the ROC curves of the three evaluated approaches: the content-based, integrated, and baseline. It is clear that the integrated approach outperforms both the content-based approach and the baseline, while the content-based approach fares worst. The difference between the integrated approach and the baseline was found to be statistically significant   ($p<0.01$ using the DeLong statistical test \cite{delong1988comparing}) as was the difference between the baseline and the content-based approach. It should be noted, though, that while the content-based approach does not use user-item data, it still did not fall far behind the other approaches that do leverage this information.

\noindent 

\vspace{5pt}

\noindent \textbf{Cold start experiments.} As described in Section \ref{subsec:experimentalSetup}, we remove a varying percentage of items from the training set (and all associated transactions) in order to create a larger percentage of cold start sessions in the test set. Table \ref{tab:dis_session_remove_items} describes the characteristics of the train and test datasets for different percentages of removed items.

\begin{figure*}
	\centering
	\begin{minipage}[b]{.4\textwidth}
	\includegraphics[height=2.3415in, width=3.024in]{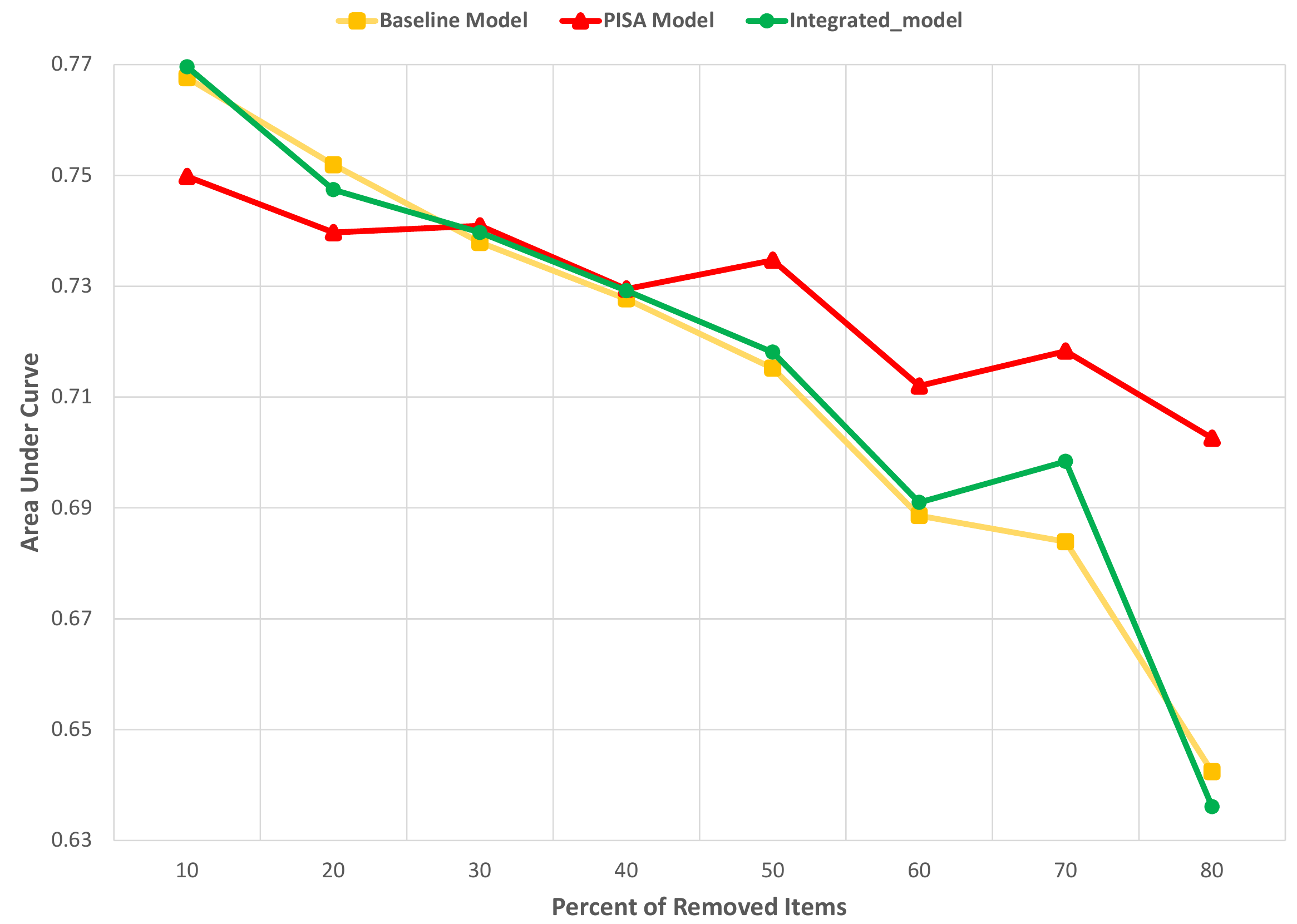}
	\caption{Cold start scenario experiments - consumption intent prediction models' AUC}
	\label{fig:remove_items_experiment}
	\end{minipage}\qquad
	\begin{minipage}[b]{.4\textwidth}
	\includegraphics[height=2.3415in, width=3.024in]{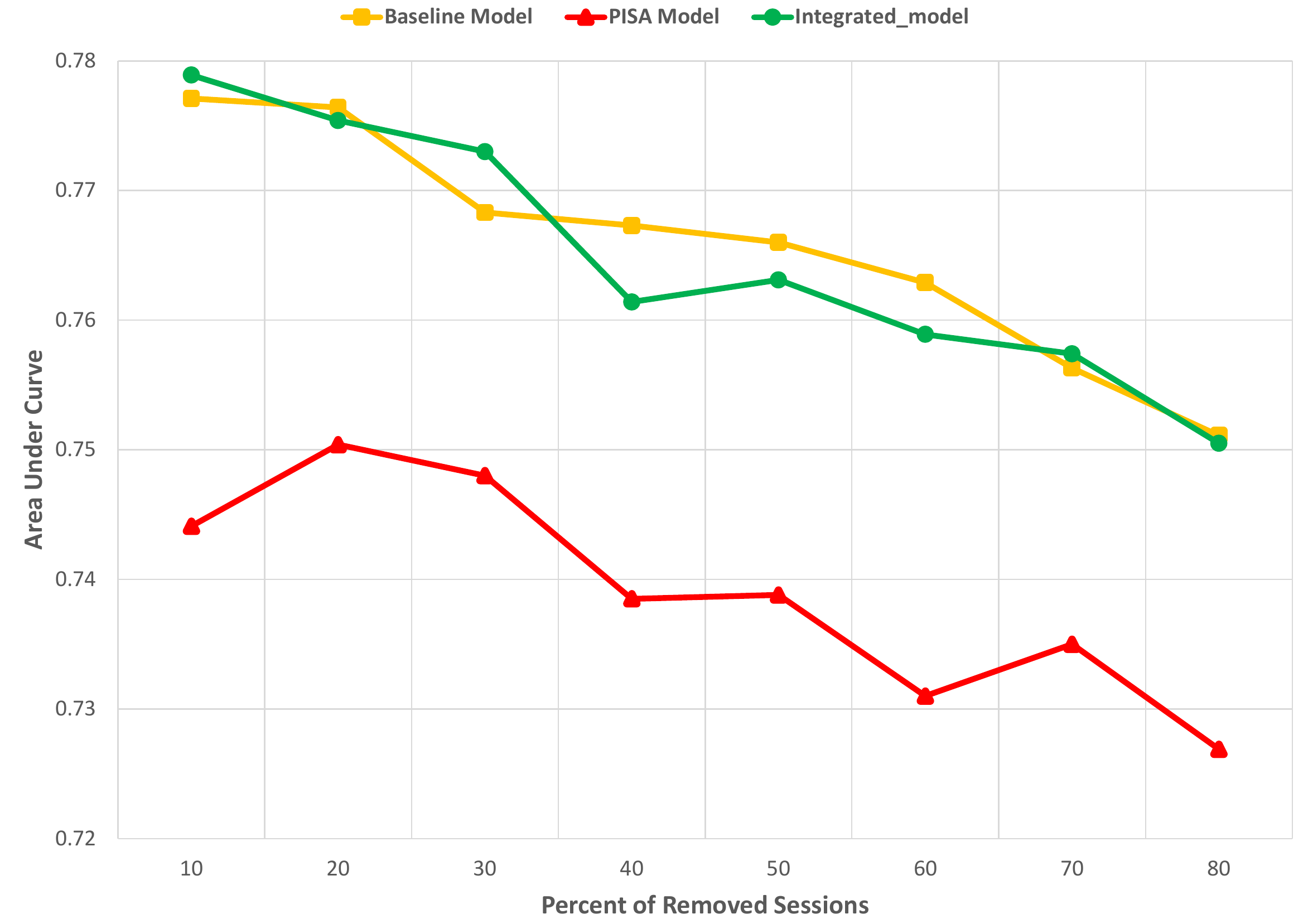}
	\caption{Random removal scenario experiments - consumption intent prediction models' AUC}
	\label{fig:remove_session_experiment}
	\end{minipage}
\end{figure*}

The results of the cold start experiments are presented in Figure \ref{fig:remove_items_experiment}. It is clear that although the content-based approach initially underperforms the other two approaches, its relative performance increases as the percentage of cold start item increases. The integrated approach also outperforms the baseline in most cases. All of the differences in performance among the three approaches are statistically significant except in the case of 30\% and 40\% removed items (the baseline and integrated approaches do not perform in a statistically significant manner in the case of 10\% and 80\% removed items).
 \newline

\begin{table*}
	\caption{Dataset statistics for the cold start experiments. The numbers in brackets are the percentage of sessions in which a ``buy'' event took place. ``Cold'' sessions are sessions which include at least one item that is not included in the train dataset. In ``warm'' sessions all of the items are included in the train dataset.} 
	\label{tab:dis_session_remove_items}
	\begin{tabular}{|c|c|c|c|c|} \hline 
		\textbf{\% Removal} & \textbf{\# Sessions -- train set} & \textbf{\# `Cold' sessions -- test set} & \textbf{\# `Warm' sessions -- test set} & \textbf{\% `Cold' sessions } \\ \hline 
		\textbf{0\%} & 1,674,964 (6.33\%) & 14 (7.14\%) & 67,740 (5.81\%)  & <0.5\%  \\ \hline 
		\textbf{10\%} & 1,382,802 (6.16\%) & 10,484 (7.02\%) & 57,269 (5.588\%) & 15.47\% \\ \hline 
		\textbf{20\%} & 1,143,907 (5.975\%) & 23,009 (6.56\%) & 44,744 (5.424\%) & 33.96\% \\ \hline 
		\textbf{30\%} & 904,386 (5.709\%) & 28,463 (6.721\%) & 39,290 (5.149\%) & 42.01\% \\ \hline 
		\textbf{40\%} & 768,666 (5.68\%) & 31,337 (7.075\%) & 36,416 (4.72\%) & 46.25\% \\ \hline 
		\textbf{50\%} & 768,666 (5.68\%) & 31,337 (7.075\%) & 36,416 (4.72\%) & 58.87\% \\ \hline 
		\textbf{60\%} & 487,472 (5.57\%) & 46,243 (6.276\%) & 21510 (4.807\%) & 68.25\% \\ \hline 
		\textbf{70\%} & 402,543 (5.972\%) & 48,365 (6.147\%) & 19,388 (4.967\%) & 71.38\% \\ \hline 
		\textbf{80\%} & 341,875 (5.708\%) &52,979 (6.227\%)&14,774 (4.312\%) & 78.19\% \\ \hline 
	\end{tabular}
\end{table*}

\noindent \textbf{Random removal experiments.} The results of this experiment are presented in Figure \ref{fig:remove_session_experiment}, and they clearly show that the content-based approach consistently underperforms compared to the other two approaches and the difference between the content-based approach and the other two approaches is statistically significant.
 
The above results verify our hypothesis that the superior performance of the content-based and integrated approaches in the cold start experiments were indeed the result of more ``cold'' items, rather than simply a smaller dataset.

\begin{table*}
  \caption{Dataset statistics for random removal experiments. The numbers in brackets are the percentage of sessions in which a ``buy'' event took place. ``Cold'' sessions are sessions which include at least one item that is not included in the train dataset. In ``warm'' sessions all of the items are included in the train dataset.}
  \label{tab:dis_session_remove_sessions}
\begin{tabular}{|c|c|c|c|c|} \hline 
\textbf{\% Removal} & \textbf{\# Sessions -- train set} & \textbf{\# `Cold' sessions -- test set} & \textbf{\# `Warm' sessions -- test set} & \textbf{\% `Cold' sessions } \\ \hline 
\textbf{0\%} & 1,674,964 (6.33\%) & 14 (7.14\%) & 67,740 (5.81\%)  & <0.5\%  \\ \hline 
\textbf{10\%} & 1,507,467 (6.34\%) & 392 (8.163\%) & 67,361 (5.796\%) & 0.58\% \\ \hline 
\textbf{20\% } & 1,339,971 (6.324\%) & 407 (7.862\%) & 67,346 (5.797\%) & 0.60\% \\ \hline 
\textbf{30\%} & 1,339,971 (6.324\%) & 407 (7.862\%) & 67,346 (5.797\%) & 0.59\% \\ \hline 
\textbf{40\%} & 1,004,978 (6.33\%) & 435 (8.506\%) & 67,318 (5.792\%)] & 0.64\% \\ \hline 
\textbf{50\%} & 837,482 (6.345\%) & 512 (8.008\%) & 67,241 (5.793\%) & 0.76\% \\ \hline 
\textbf{60\%} & 669,986 (6.327\%) & 593 (7.757\%) & 67,160 (5.792\%) & 0.88\% \\ \hline 
\textbf{70\%} & 502,489 (6.335\%) & 800 (6.75\%) & 66,953 (5.798\%) & 1.18\% \\ \hline 
\textbf{80\%} & 334,993 (6.405\%) & 1,074 (6.238\%) & 66,679 (5.802\%)
 & 1.59\% \\ \hline 
\end{tabular}
\end{table*}

\section{Discussion}
\noindent \textbf{Additional analysis of cold start scenarios.} In order to further analyze the performance of our model in cold start scenarios, we conducted an additional set of tests. We trained our models on the original train set but evaluated it on two test sets with varying percentages of cold start sessions. The first test set contained \textit{no cold-start sessions}, while in the other test we set the ratio of the two to 50\%.

\begin{figure*}
	\centering
	\begin{minipage}[b]{.4\textwidth}
		\includegraphics[height=2.3415in, width=3.024in]{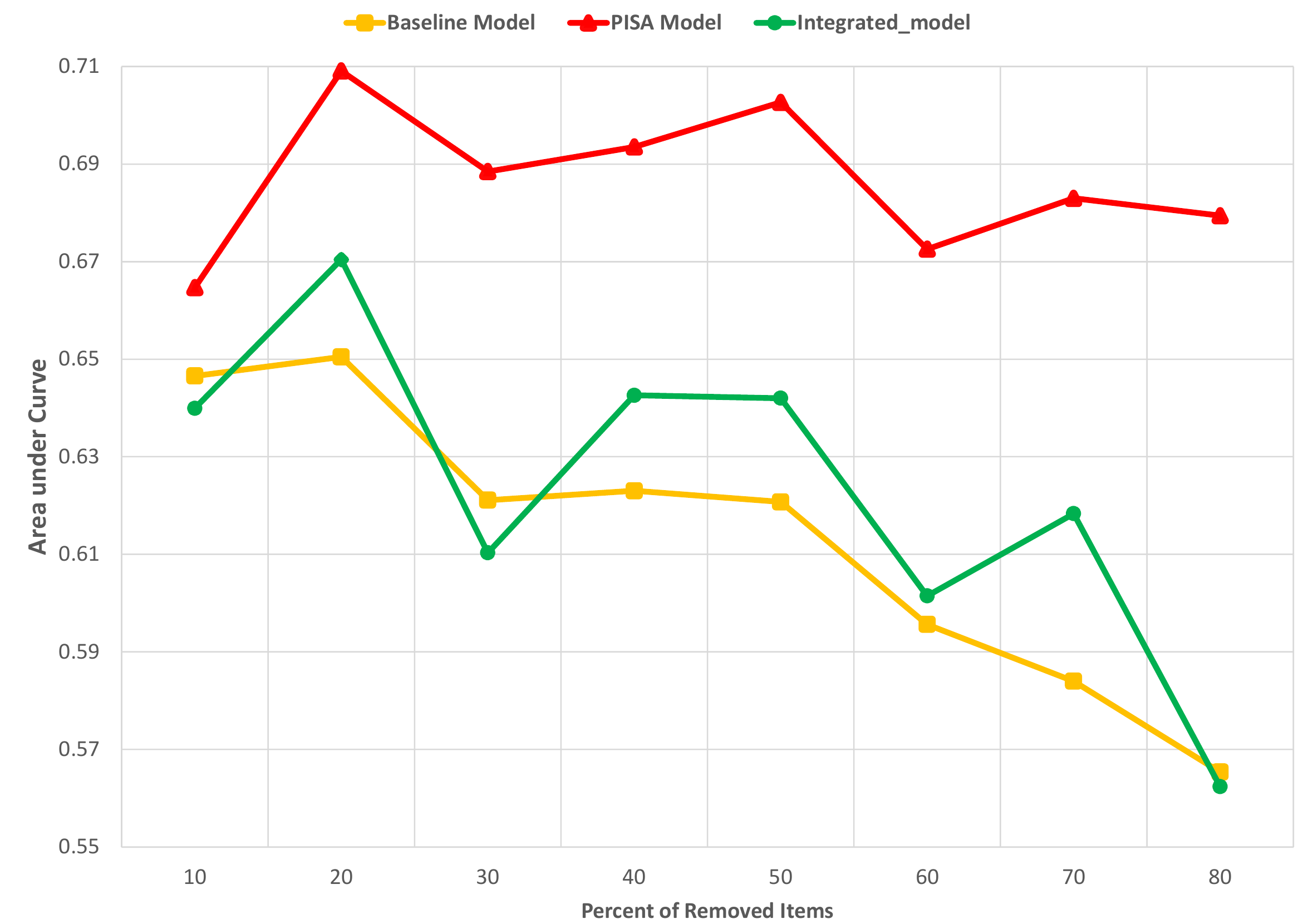}
		\caption{Cold start scenario experiments - consumption intent prediction average precision (at least 50\% cold items per session)}
		\label{fig:75_new_items_avg_prec_recall}
	\end{minipage}\qquad
	\begin{minipage}[b]{.4\textwidth}
		\includegraphics[height=2.3415in, width=3.024in]{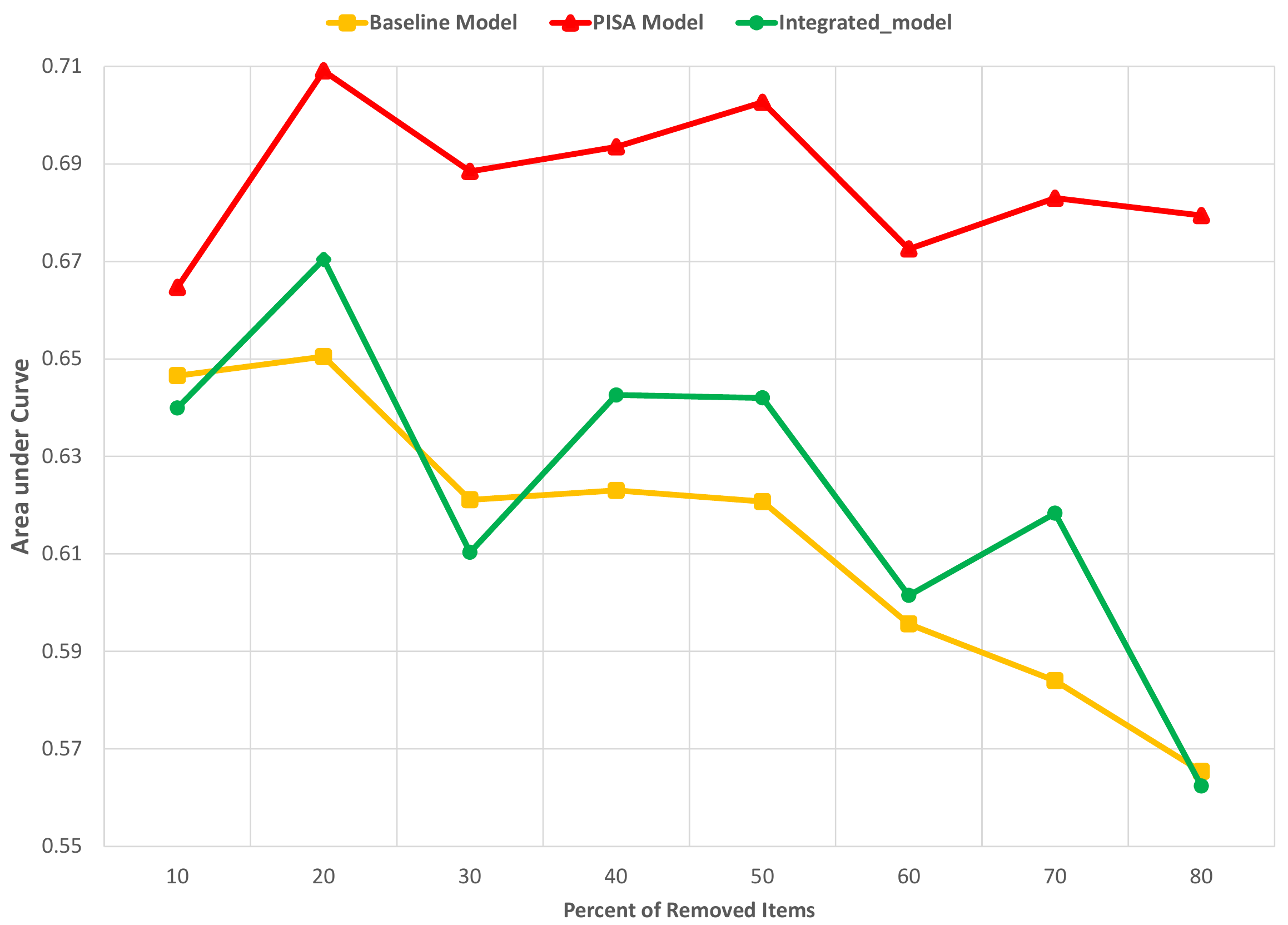}
		\caption{Cold start scenario experiments - consumption intent prediction AUC (at least 50\% cold items per session)}
		\label{fig:75_new_items}
	\end{minipage}
\end{figure*}

\begin{table*}
	\caption{Comparison between categories} 
	\label{tab:category_des}
	\begin{tabular}{|c|c|c|c|c|} \hline 
		\textbf{attribute} & \multicolumn{2}{c|}{\textbf{categories for which content-based outperformed}	} & \multicolumn{2}{c|}{\textbf{categories for which baseline outperformed}}  \\ \hline 
		& Average & Standard Deviation & Average & Standard Deviation \\ \hline 
		Number of items & 9,104	& 4,109 & 25,444 & 44,933 \\ \hline 
		Item description length (\# of words) & 11.23 & 0.9 & 8.15 & 3.44 \\ \hline 
		Number of Sessions & 1534 & 1611 & 6349 & 7416 \\ \hline

	\end{tabular}
\end{table*}

\begin{figure}[H]
	\includegraphics[height=2.1324375in, width=2.754
	in]{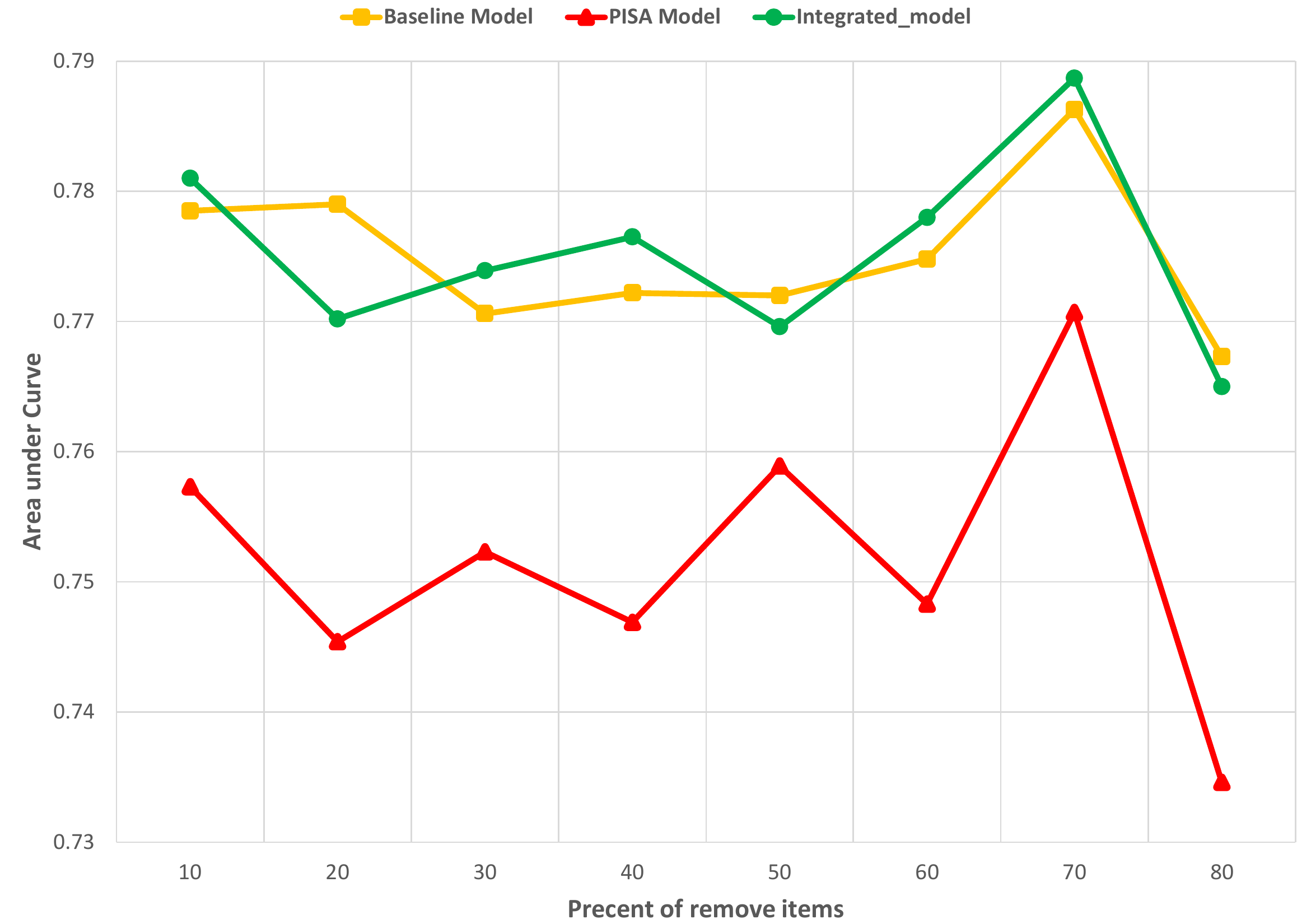}
	\caption{Cold start scenario experiments -- consumption intent prediction AUC for sessions with no cold items}
	\label{fig:no_new_items}
\end{figure}

Figures \ref{fig:no_new_items}, \ref{fig:75_new_items} and \ref{fig:75_new_items_avg_prec_recall} present our results for the two test sets. In Figure \ref{fig:no_new_items}, we can see that the content-based approach fares significantly worse when no cold start items are included in the sessions. It should be noted, though, that the integrated approach outperforms the baseline (the difference was not statistically significant). On the other hand, when the percentage of sessions with cold start items was 50\% (Figures \ref{fig:75_new_items} and \ref{fig:75_new_items_avg_prec_recall}), the content-based approach outperformed the integrated approach and the baseline by a wide margin. It is also important to note that the integrated approach outperforms the baseline in this scenario as well.
\newline

\begin{figure*}
	\includegraphics[height=3.127575
	in, width=4.0392
	in]{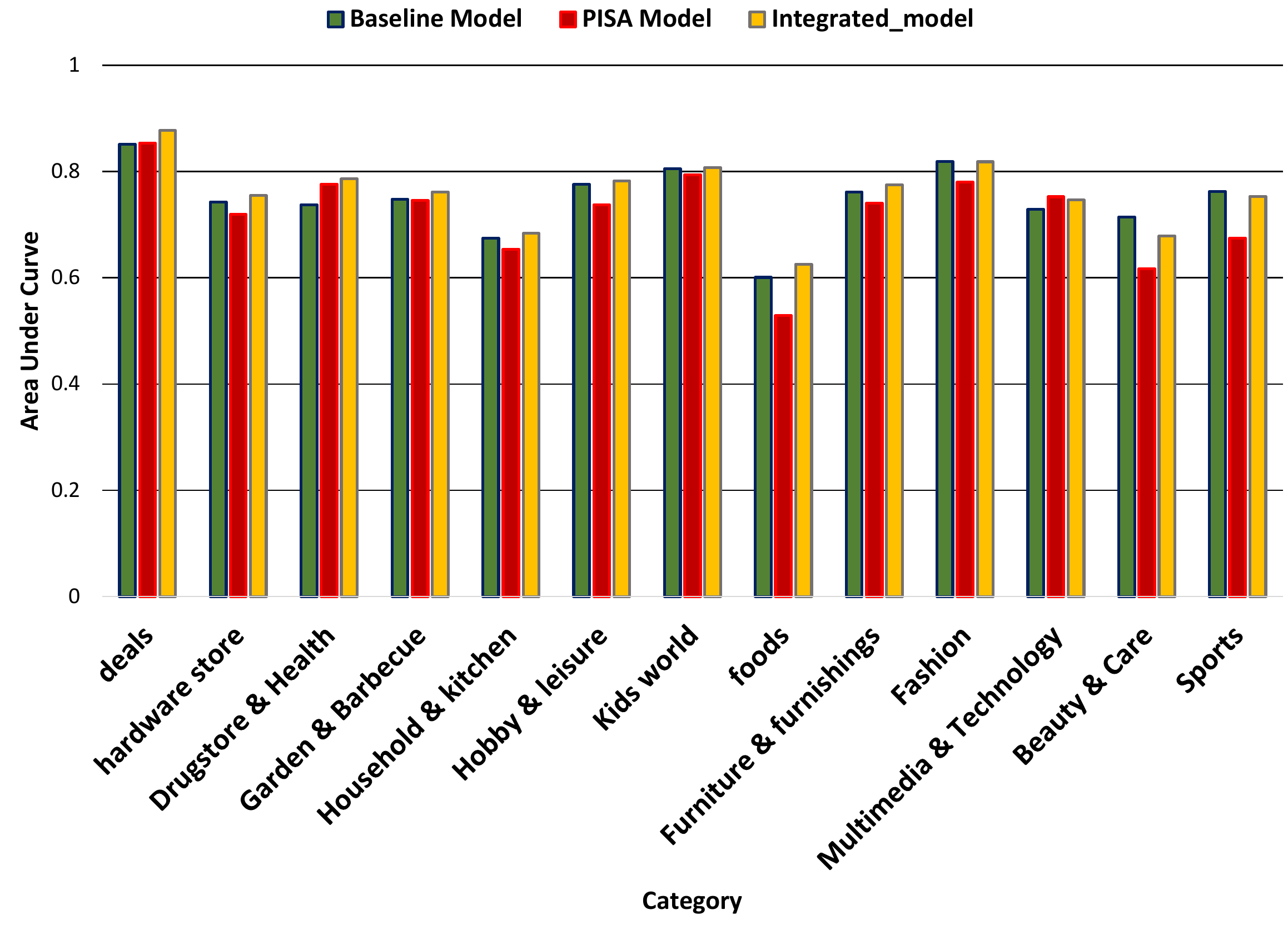}
	\caption{Consumption intent prediction AUC by category}
	\label{fig:regular_category}
\end{figure*}

\noindent \textbf{Evaluating our approach across different product categories.} While the content-based model significantly outperforms the baseline when the number of cold start sessions is high, we wanted to further understand the conditions that enable our model to outperform the baseline under "normal" circumstances. For this reason, we conduct a category-by-category analysis of the results of the "All-Data" experiments (see Section \ref{subsec:experimentalSetup}). Our analysis found major differences in the performance of our model across the different product categories, as shown in Figure \ref{fig:regular_category}. While it is difficult to reach definitive conclusions due to the black-box nature of deep neural networks, several insights can be drawn:
\begin{itemize}
\item \textbf{Sufficient number of items.} As in many applications involving deep learning, a sufficient amount of data is needed to ensure that the deep neural architecture converges. This is apparently the case not only at the dataset level – where we had >250,000 items to learn from despite not all of them being included in the sessions - but at the category level as well. In general, our approach fared better in categories where the number of items was high.
It is important to note that while the average number of items-per-category presented in Table \ref{tab:category_des} may lead to the opposite conclusion, this value is skewed due to a small number of categories with an order of magnitude more items than others, despite never being part of a session. The high standard deviation for these product categories illustrates this point.

\item \textbf{Item description length.} While our experiments show that even a short description of each item is sufficient in order for \MethodName to perform, our analysis verifies the rather intuitive conclusion that longer item descriptions enable our approach to perform better. As shown in Table \ref{tab:category_des}, the product categories in which our method outperformed the baseline tend to have longer descriptions (on average) and a smaller standard deviation.

\item \textbf{Number of training sessions.} Somewhat surprisingly, the number of sessions available for training had no discernible effect on our model's performance. In fact, some of our best-performing categories had relatively few sessions, as shown in Table \ref{tab:category_des}. This leads us to conclude that our approach is capable of learning across categories and that our word embeddings are effective in modeling the latent connections among items.
\end{itemize}

\noindent We also try to assess the influence of the number of clicked items in sessions. Unfortunately, there are no significant differences between the models. We believe that this is the result of the diversity among the users of the e-commerce website.

\section{Conclusions and Future work}

In this study, we present \MethodName, a content-based approach for the cold start scenario in session-based purchase prediction. Our approach uses word embeddings to model the content of items from multiple categories and provides these embeddings as input to a recurrent neural network. \MethodName  is highly effective in cold start scenarios, where multiple items are not previously known, but it is less effective in small datasets. We believe that using content data can be useful in sessions that have new items. Our approach is also highly effective when combined with standard user-item recommendation systems; our evaluation shows that \MethodName outperforms the other approaches and that the difference between \MethodName's performance and the performance of the other approaches is statistically significant in many cases.

For future work, we plan to extend our framework so it can provide purchase prediction for specific items in a session. In addition, we plan to test additional deep architectures - such as autoencoders. for the content-based recommendation process. Finally, we intend to enrich our input with meta-data such as timestamps, click counts, geographical location, etc.

\clearpage
\clearpage
\bibliographystyle{ACM-Reference-Format}
\bibliography{sample-bibliography}

\end{document}